%% file: main.tex
\documentclass[dvipsnames,format=sigconf, nonacm]{acmart}

\AtBeginDocument{%
  \providecommand\BibTeX{{%
    \normalfont B\kern-0.5em{\scshape i\kern-0.25em b}\kern-0.8em\TeX}}}

\usepackage{booktabs} %

\usepackage{tikz}
\usetikzlibrary{arrows,calc}
\usepackage[table]{colortbl}%

\usepackage{subcaption}
\usepackage{qtree}
\usepackage{xspace}
\usepackage{multirow}
\usepackage{listings}   
\usepackage{enumitem}
\usepackage{float}
\usepackage{setspace}

\begin{document}
\title{Learning Semantics-aware Search Operators for Genetic Programming}

\author{Piotr Wyrwiński}
\orcid{0000-0001-9796-5025}
\affiliation{\institution{Poznan University of Technology}
 \city{Poznan}
 \country{Poland}
}
\authornote{Corresponding author: piotr.wyrwinski@cs.put.poznan.pl}

\author{Krzysztof Krawiec}
\orcid{0000-0001-5439-3231}
\affiliation{%
 \institution{Poznan University of Technology}
 \city{Poznan}
 \country{Poland}
}

\renewcommand{\shortauthors}{}

\newcommand{\mnamens}{NEON} %
\newcommand{\mname}{\mnamens\xspace}

\begin{abstract}
Fitness landscapes in test-based program synthesis are known to be extremely rugged, with even minimal modifications of programs often leading to fundamental changes in their behavior and, consequently, fitness values. Relying on fitness as the only guidance in iterative search algorithms like genetic programming is thus unnecessarily limiting, especially when combined with purely syntactic search operators that are agnostic about their impact on program behavior. In this study, we propose a semantics-aware search operator that steers the search towards candidate programs that are valuable not only actually (high fitness) but also only potentially, i.e. are likely to be turned into high-quality solutions even if their current fitness is low. The key component of the method is a graph neural network that learns to model the interactions between program instructions and processed data, and produces a saliency map over graph nodes that represents possible search decisions. When applied to a suite of symbolic regression benchmarks, the proposed method outperforms conventional tree-based genetic programming and the ablated variant of the method.  

\end{abstract}

\begin{CCSXML}
<ccs2012>
   <concept>
       <concept_id>10010147.10010148</concept_id>
       <concept_desc>Computing methodologies~Symbolic and algebraic manipulation</concept_desc>
       <concept_significance>500</concept_significance>
       </concept>
   <concept>
       <concept_id>10010147.10010257</concept_id>
       <concept_desc>Computing methodologies~Machine learning</concept_desc>
       <concept_significance>500</concept_significance>
       </concept>
 </ccs2012>
\end{CCSXML}

\ccsdesc[500]{Computing methodologies~Symbolic and algebraic manipulation}
\ccsdesc[500]{Computing methodologies~Machine learning}

\ccsdesc[500]{Computing methodologies~Machine learning approaches}

\keywords{genetic programming, symbolic regression, graph neural networks}

\maketitle
\input{body}

\bibliographystyle{ACM-Reference-Format}
\bibliography{bibliography} 

\end{document}

%% file: body.tex
\section{Introduction}

One of the main challenges in developing search heuristics consists in designing effective search operators. In evolutionary computation, there are broadly speaking two alternative practices that have crystallized over years. One of them relies on generic search operators that are not tailored to the specifics of a particular domain, problem, or solution representation. Random bit-flip mutations and point crossovers used in basic genetic algorithms are good examples of this category. The alternative practice consists in designing operators that are more bespoke for solution representation or/and domain, and attempt to exploit its properties for the sake of search effectiveness. In tree-based genetic programming (GP) that is the subject of this study, this may assume various forms and extents. For instance, biasing a tree-swapping crossover towards inner tree nodes (and away from leaves), is a simple means of `nudging' the search operator so as to reduce the bloat and avoid excess of dead code (introns). Another example is the homologous tree-swapping crossover, meant to preserve the building blocks and their `anchoring' to specific locations in the tree blueprint. Last but not least, the semantic-aware search operators attempt to improve the behavior of the offspring solutions on the training data.  

Many of the domain-aware search operators have been indeed successful at improving the efficiency and efficacy of evolutionary search. However, thy were manually designed and thus potentially suboptimal. In particular, they may be burdened by unconscious biases of the designer, or have some side-effects that undermine search efficiency in some (not always anticipated) way. For instance, the exact geometric semantic GP operators offer appealing convergence properties, but lead to exponential growth of program size. More importantly, however, given the virtually unlimited design space of search operators, we can be sure that those proposed by humans have barely scratched the surface: there is a plethora of hypothetical search operators that researchers did not come up with yet for all sorts of reasons, like conceptual complexity, difficulty of technical implementation, or even proverbial lack of imagination. 

Given these observations and the impressive progress within contemporary machine learning (ML), \emph{in this study we pose the problem of designing search operators for GP as a learning/optimization problem}. Rather than burdening a human expert with this task, we attempt to learn a crossover operator that anticipates the long-term effects of appointing the offspring solution from a range of possible options. More specifically, given a parent candidate program, our operator learns how to modify it in a way that is likely to prospectively, after a number of further modifications, translate into synthesizing the correct, final solution. This informed choice-making provides additional guidance for the search process, in contrast to conventional evolutionary algorithm that relies only on fitness function to `probe' the candidate solutions. The proposed method, dubbed \textbf{NEural cOmbiNer (\mname)}, forms the main contribution of this study.

In the following Sec.\ \ref{sec:approach}, we provide the rationale for \mname's design and detail its operation, including the graph neural network (GNN) it relies on. In Sec.\ \ref{sec:related}, we contextualize our approach by linking it to the prior work. Section \ref{sec:experiments} presents the results of extensive computational experiment involving almost a hundred of symbolic regression benchmarks. Sections \ref{sec:discuss} and \ref{sec:conclusions} conclude the paper with discussion of results and presentation of the broader context.

\section{Proposed approach}\label{sec:approach}

\subsection{Motivations}\label{sec:motivations}

We devised \mname  with two primary goals in mind. The first one originates in the observation that, similarly to other heuristics, EC algorithms perform \emph{iterative, incremental search} and as such are not expected to solve a problem in a single step. This implies that search operators should be aligned with the incremental nature of search, and in general \textbf{care more about the long-term than immediate effects}. Indeed, this is the main feature that distinguishes advanced heuristics from the greedy local search. Achieving such capacity with manual design is particularly difficult. Therefore, we delegate this task to an ML model, which predicts which of a number of possible modifications of the parent program is most likely to translate into relative improvement of its descendant solutions -- in principle not only the direct offspring, but descendants in multiple subsequent generations.   

The second goal stems from the observation that in typical variants of GP, search operators are generic and abstract from the specifics of the \emph{domain-specific language} (DSL) used for representing candidate solutions (programs). For instance, the conventional tree-swapping crossover is commonly used for symbolic regression (SR), boolean function synthesis, construction of classifiers, evolution of problem-solving agents, and more. Even more so, search operators are not only agnostic about the \textit{class} of problems, but also about the specific \textit{instance} of a problem. For instance, a search operator tailored to SR is typically not explicitly informed about the goal of search, i.e. the desired behavior (output) of the program for individual training examples (fitness cases). Semantic-aware search operators are rare exceptions from this rule. 

Therefore, our second design choice is to \textbf{inform \mname  about selected characteristics of the domain and problem instance}. Concerning the problem instance, \mname has access to the entire training data that forms a SR task, i.e. set of examples (fitness cases), each comprising a vector of input (independent) variables and the corresponding scalar values of the dependent variable (desired output). %
When it comes to the domain, the method is `aware' about the operational/semantic characteristics of individual instruction of the DSL, i.e., in the case of SR, of the mathematical operations and functions. We achieve that by equipping it with a GNN that is trained by observing the results of applying instructions to various arguments, as detailed in the following.

\subsection{\mname algorithm}

\mname is a ML-based search operator for tree-based GP. In the experimental part of this paper, we apply it to SR problems, so the description that follows is illustrated with examples from this domain; nevertheless, it can be applied also to other domains and DSLs. 

In SR, the task is to construct a mathematical expression that maps a number of independent variables $x_i$ to a dependent variable $y$ so that the regression \emph{model} obtained in this way minimizes an approximation error (typically MSE) on a set of training examples $T=\{(\mathbf{x}^{(j)},y^{(j)})\}$, and prospectively generalizes well beyond this sample. SR is a special case of program synthesis from examples, where the search space is defined by the DSL comprising the set of instructions $O$ (mathematical operators) and the set of terminals $V$, i.e. input variables $x_i$ and constants. Any finite tree formed by composing the elements of $O$ and $V$ is a valid program. 

\mname comprises three main components, detailed in the following:
\begin{itemize}
    \item The \textbf{library} of programs,
    \item The \textbf{expansion} algorithm,
    \item The \textbf{grafting} search operator.
\end{itemize}

\subsubsection{The library}\label{sec:library}

The centerpiece of the proposed method is a library $L$, realized as a FIFO queue. $L$ is initially empty, and is intended to store the potentially valuable programs created by the method in an evolutionary run, independently from the main population $P$. Newly created programs are enqueued to $L$ until its length becomes equal to the population size. From then on, enqueuing of a next subprogram is accompanied with dequeuing of a subprogram from the other end of the queue, so that $L$'s length never exceeds the population size. Therefore, $L$ always holds the most recent subprograms, while the older ones are systematically discarded. 

The library is supplied with new subprograms by the GNN-based expansion algorithm described in the following. The collected subprograms are then used by a grafting operator presented in Sec.~\ref{sec:grafting}.

\subsubsection{The expansion algorithm}\label{sec:expander}

\begin{figure}[t]
    \centering
    \includegraphics[width=0.7\linewidth]{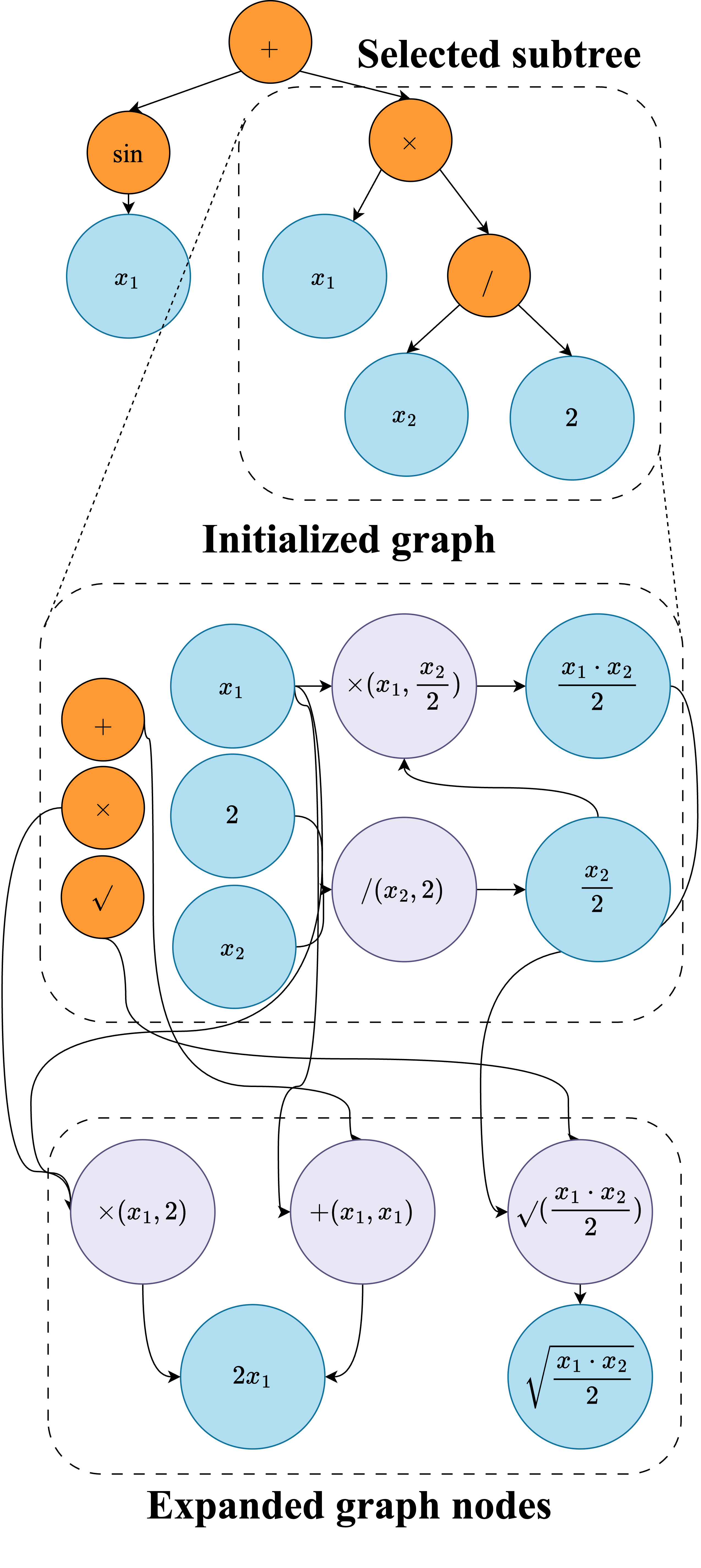}%
    \caption{An example of graph expansion performed by \mname. The subtree $s$ selected from the source program is converted to a graph representation. The graph is then expanded by a layer of nodes that combines $s$ with the available operations and subtrees of $s$. Then, the GNN is queried on the expanded graph and decorates the expanded nodes with saliency values, which are used to appoint the programs to be added to the library (see Sec.\ \ref{sec:expander}). }
    \label{fig:graph-example}
\end{figure}

The role of the expansion algorithm (\textbf{expander} for short) is to form potentially valuable subprograms from the pieces of code available in the population. Given a population $P$, the expander operates as follows:
\begin{enumerate}
    \item \textbf{Draws} at random a fraction of candidate solutions (program trees) from $P$, without replacement, and stores them in the set $S$. The drawing is intended to reduce the overall computational cost incurred by the steps that follow. In the experiments conducted in this paper, 20\% of solutions were drawn from $P$. 
    \item For each program tree $p \in S$, \textbf{samples} at random a node in $p$ and extracts from it the subtree (subprogram) $s$ rooted at that node. Rather than sampling uniformly from $p$ (which would result in exponential overrepresentation of shallow subprograms), the sampling is biased to prefer higher (larger) subprograms in the following way: 
    \begin{enumerate}
        \item Let $h_p$ be the height of $p$.  
        \item A height $h$ is drawn proportionally to the normalized distribution $[1, 2, \ldots, h_p]$.  
        \item From all subtrees of $p$ of height $h$, one is drawn at random. 
    \end{enumerate}
    \item \textbf{Expands} $s$ in a number of ways at its root node, by combining it exhaustively with the terminals from $V$ and all subtrees of $s$. Each expansion involves `topping' $s$ with a single instruction $o$ from $O$ and providing $o$ with the remaining arguments, increasing so the height of $s$ by one. The result of this step is a set of candidate programs $C$ that all share $s$ as one of their subtrees and so form a single \emph{graph} $G$, detailed in Sec.\ \ref{sec:gnn-sel}. 
    \item \textbf{Selects} up to $n$ programs in $C$ and adds (enqueues) them in $L$. The selection is based on the GNN-based mechanism presented in Sec.\ \ref{sec:gnn-sel}.  
\end{enumerate}
Steps 2-4 are repeated for each drawn program $p$ independently, and are depicted in Fig.\ \ref{fig:graph-example}.

In experiments, we set $n=5$. Given that the expander processes 20\% of the population, the above procedure has the chance to replenish the entire library $L$ with new programs (as the capacity of $L$ is $|P|$). However, if the selection mechanism decides to send less than $n$ programs derived from some $p$s to $L$, this will not be the case. This implies that $L$ will be usually dominated by programs built in the current generation, but it may also hold some programs from the previous generations (typically only from the previous one).   

Importantly, the programs in $C$ are \emph{not} evaluated using fitness function: the selection mechanism is meant to identify the programs that are \emph{prospectively} most promising for the further course of search (as argued at the beginning of Sec.\ \ref{sec:approach}), i.e. as potential subprograms to be inserted into programs by the grafting operator described in the following.  

\subsubsection{Grafting operator}\label{sec:grafting}
The grafting operator implants the programs collected in $L$ into the parent programs selected from the current population $P$ (using some adopted selection algorithm, e.g. tournament selection). It is implemented like a typical evolutionary search operator, and can be thus used alongside (e.g. chained with) conventional evolutionary search operators like mutation or crossover.  

Our implementation of grafting resembles a tree-swapping crossover with a program from $L$. Given a parent program $p$, it proceeds as follows:
\begin{enumerate}
    \item Draws a node $s$ in $p$, using an adopted node-sampling scheme (uniform sampling in the experiments reported in this paper). 
    \item Draws at random a program $p_l$ from $L$ (uniformly). 
    \item Replaces in $p$ the subtree rooted in $s$ with $p_l$.  
\end{enumerate}
As a result, the program $p_l$ fetched from $L$ becomes a \emph{subprogram} of the offspring solution. Importantly, $p_l$ has been earlier created by the scanner via one-level expansion of another program, and appointed as potentially valuable by a trained GNN, informed by the properties of the given problem instance. This design choice is consistent with the arguments brought up at the beginning of Sec.\ \ref{sec:approach}, i.e. that search operators should be incremental and informed about the problem being solved.

\subsection{GNN-based selection mechanism}\label{sec:gnn-sel}

\mname employs a GNN to make informed choices about the expected usefulness of programs created in by the expansion step (Sec.\ \ref{sec:expander}). The adopted graph representation is more sophisticated than GP trees (features more node types and more nodes that the GP tree it has been built from), so we detail it in the following. This process is agnostic about the the architecture of the GNN, which is covered in Sec.\ \ref{sec:gnn-arch}.

Let us emphasize that one of the main motivations of using a GNN, rather than some other type of neural architecture (e.g., one of the contemporarily popular transformers) is that the solutions produced in the process are, as it will be shown in the following, \emph{syntactically correct by design}. This is so because the graph nodes follow the rules of the grammar of the underlying DSL. Arguably, this does not buy one much when considering SR problems, where all expressions have the same type and the only syntactic constraint is the arity of operators in $O$. Nevertheless, for DSLs with richer type systems, this characteristic would be advantageous.

\subsubsection{Graph representation}

The intent behind the method is to learn to reason about the semantic effects of applying DSL instructions to existing partial solutions (subtrees). Therefore, the graph $G=(N,E)$ presented to the GNN is more verbose than GP solutions in the population. In particular, it features two types of nodes:
\begin{itemize}
    \item \emph{application nodes} $N_a$, each representing the application of an operation from $O$ to concrete arguments, 
    \item \emph{value nodes} $N_v$, which represent the outcomes of those applications, including their semantic effects (in terms of the values calculated for individual training examples). 
\end{itemize}
As an example, performing the $\times$ operation on the variable and constant nodes $x_1$ and $2$ creates an application node labeled $\times(x_1,2)$, along with a separate value node storing the computed result, $2x_1$. In the case of a dataset containing $n$ examples, these scalar values are substituted with $n$-dimensional vectors assigned to value nodes. Similarly, both initial constants and variables are represented as value nodes (Fig.~\ref{fig:graph-example}).

The causal relationships among the nodes in $N = N_a \cup N_v$ are represented by directed edges (arc) in $E$. Each application node receives one incoming edge from an operation node ($\in O$), along with $k \geq 1$ incoming edges from variable, constant, or pre-existing value nodes ($\in X \cup C \cup V$), where $k$ corresponds to the arity of the operation. Additionally, each application node has a single outgoing edge that connects to the value node representing the operation's result ($\in N_a \times V$).

\subsubsection{Graph expansion}\label{sec:graph_expansion}

Before undergoing expansion (step 3 of the expander, Sec.\ \ref{sec:expander}), the subtree expression $s$ is converted to the above form by parsing it bottom-up from leaves to the root, creating the corresponding value nodes and application nodes, and chaining them accordingly with graph arcs.
If a newly created application node produces a value $v$ that already exists in the graph, it is directly linked to the corresponding value node $v$ (see application nodes $+(x_1, x_1)$ and $\times(2, x_1)$ in Fig.~\ref{fig:graph-example}). Otherwise, a new value node $v'$ is introduced. The equivalence between $v$ and $v'$ is established through symbolic execution: we traverse the edges connecting these nodes back to the graph's initial nodes, constructing the symbolic expressions $v'(\mathbf{x})$ and $v(\mathbf{x})$. To determine whether they are identical, we check if $v'(\mathbf{x}) - v(\mathbf{x}) \equiv 0$ using a symbolic executor from the SymPy library~\cite{10.7717/peerj-cs.103}. This approach ensures that the graph remains \emph{minimal} by avoiding redundant value nodes.

Once $s$ is represented as a graph, the actual expansion follows. It consists in two steps:
\begin{itemize}
    \item Adding a single `layer' of application nodes, representing all possible applications of the DSL operations in $O$ applied to all arguments available in the graph. Each application node added in this way embodies an application of a specific instruction/operator $o \in O$ to a unique combination the DSL terminals from $V$ and all the value nodes in the graph representation of $s$. 
    \item Calculating the values represented by the newly added application nodes, and creating the corresponding value nodes. Similarly to the conversion of $s$ to graph representation described above, no duplicate value nodes are allowed. If the value node about to be created is already present in the graph, the outgoing arc of the application node is connected to it, rather than creating a new value node. 
\end{itemize}
This process results in an expanded graph $G$, which is used to query the GNN.

\subsubsection{Search guidance with the GNN} 

Once the graph $G$ has been expanded, the GNN functions as an \emph{attention mechanism}, identifying the graph nodes that correspond to potentially valuable partial solutions (subprograms). More precisely, it generates a \emph{saliency map} over the set of application nodes 
$N_a$ in G, effectively mapping the graph to a probability distribution: 
$G \mapsto (0,1)^{|N_a|}$. %

To accommodate multiple examples within the dataset 
$T$ associated with the SR problem, the GNN is applied separately to each of them. 
Specifically, for the $j$th data point $(\mathbf{x}^{(j)},y^{(j)})$ from $T$, we \emph{instantiate} the graph $G$ by assigning the input values $x_i^{(j)}$s to the corresponding variable nodes in $V$. The dependent values at all value nodes in $G$ are then computed\footnote{This is efficiently managed using a cache that persists across iterations, minimizing computational overhead.}  the dependent values in all value nodes in $G$.
With this instantiated input, the GNN processes the graph and generates a saliency map $s_j$ over application nodes. The final saliency map $s$ is obtained by averaging $s_j$s across all examples in $T$. 

To determine the most promising expressions or programs (step 4 of the expander, Sec.\ \ref{sec:expander}), selection is carried out in two stages.  First, application nodes with $s_j< 0.5$ are discarded, as they are deemed unlikely candidates for expansion. From the remaining nodes, the top $k=5$ nodes with the highest saliency values are chosen, where $k$ is a configurable parameter\footnote{While alternative selection strategies were tested, this approach yielded the best results.}.

\subsection{The Graph Neural Network}

\subsubsection{Presentation of graph for the GNN}
A node in the graph, instantiated for the $j$th data point, is represented as a vector input to the GNN with the following components:
\begin{itemize}
    \item A one-hot encoded categorical feature specifying the node type (variable, value, operation, or application), contributing 4 dimensions to the vector.
    \item For operator nodes, a one-hot encoded index indicating the corresponding operator from the DSL, spanning 11 dimensions (one for each of the 11 operators used in the experiments).
    \item For value nodes:
    \begin{itemize}
        \item An embedding of the instantiated numerical value, using a 32-bit representation similar to the approach proposed in \cite{Kamienny_d’Ascoli_Lample_Charton_2022}. This encoding treats each bit of the significand and exponent in the IEEE-754 single-precision floating-point format as a distinct input to the model, contributing 32 dimensions. For other node types, this embedding is set to zero.
        \item A representation of the signed difference between the node's value and the target value $y^{(j)}$, encoded in the same manner, adding another 32 dimensions.
    \end{itemize}
\end{itemize}
Thus, the total dimensionality of the node representation vector is $79$.

\subsubsection{GNN architecture}\label{sec:gnn-arch}

\begin{figure*}[t]
    \begin{center}
    \includegraphics[width=\linewidth]{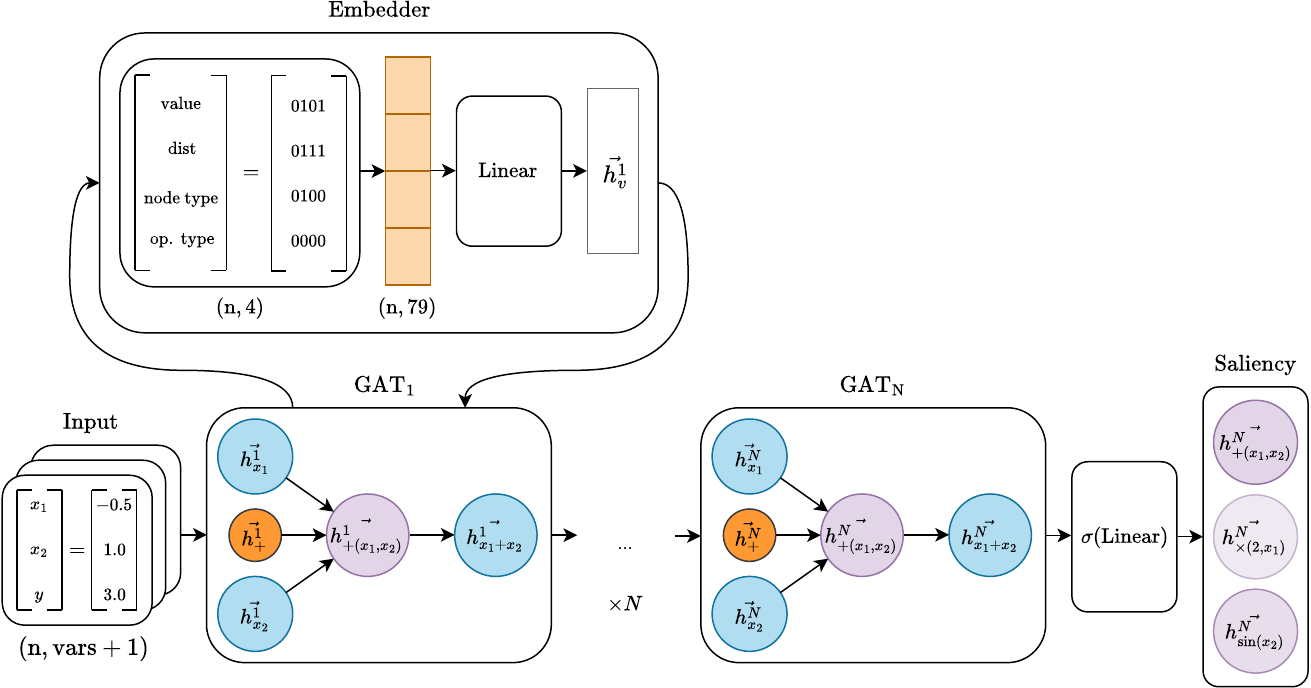}%
    \end{center}
    \caption{The architecture of the GNN used for saliency estimation in \mname; $n$: the number of examples in the dataset that specifies the SR problem instance; $N$: the number of message passing iterations and of the GAT layers of the model.}
    \label{fig:gnn}
\end{figure*}

We develop a bespoke GNN architecture inspired by the Graph Attention Network (GAT) framework~\cite{velivckovic2018graph}. GATs are well-suited for analyzing graph-structured data as they dynamically assign attention weights to neighboring nodes during multiple rounds of \emph{message passing} (described below). This mechanism enables the model to prioritize essential information, thereby capturing complex dependencies within the graph structure (Fig.\ \ref{fig:gnn}).

Similar to most GNNs, GAT maintains a \emph{state} variable $h_v$ for each node $v$ in the input graph $G$. This state is initialized with information specific to $v$, updated iteratively through message passing, and ultimately used to compute the node's final output representation. The initial state $h_v$ is obtained by applying a linear transformation to the 79-dimensional node representation vector (described earlier), mapping it to a 256-dimensional space.

The message-passing process is carried out using three stacked GAT layers, each performing one iteration of state updates for all nodes in the graph. Each GAT layer employs four independent attention heads, where each head processes the 256-dimensional node states and generates a 64-dimensional output. These outputs from all heads are concatenated, preserving the overall state dimensionality of 256, before being passed to the next layer. To aggregate incoming messages from neighboring nodes, we utilize the sum aggregation operator. Additionally, the Exponential Linear Unit (ELU) activation function~\cite{clevert2015fast} is applied in each GAT module. Throughout this process, nodes exchange information with their local neighborhood, allowing the model to iteratively refine their representations based on contextual cues.

After message passing concludes, the final node states $h_v$ are processed through an output layer, applied independently to each node in the graph. This layer consists of a single unit with a sigmoid activation function, which condenses the learned information from the message-passing stages into a binary classification decision for each node. This classification step is integral to the training procedure described in the following sections.

\subsubsection{GNN Training}\label{sec:training}
We train the GNN on a dataset of SR expression trees sampled from a predefined DSL, with tree heights of up to six in our experiments. The training process is conducted in a supervised manner, where the model learns to identify relevant subexpressions for graph expansion.
For each sampled expression tree, we randomly select a depth from $[1, \mathrm{max\_tree\_depth}]$ at which to split it, dividing the tree into two parts:
\begin{itemize}
    \item The upper part, always containing the root node,  includes the nodes we need to reconstruct. %
    \item The lower part consists of subtrees (subprograms) that were originally part of the full expression tree.
\end{itemize}
Subtrees from the lower part are placed into the graph, and a single step of graph expansion is performed (as described in Sec.\ \ref{sec:graph_expansion}). This process generates a new graph where the subtrees (subprograms) increase in height by one level. 

The newly expanded subprograms, referred to as the \textit{front}, are then sought in the upper part of the tree. Based on this comparison, we construct a binary target saliency vector: newly generated node is assigned to the positive class if it exists in the upper part of the original tree. Otherwise, it is assigned to the negative class.
The GNN's output layer predicts saliency values for the expanded nodes, which are then optimized using a binary cross-entropy loss function by comparing them with the target saliency vector. This training procedure ensures that the model learns to prioritize the most relevant subexpressions within a single expansion step.
More details on this process are given in Sec.\ \ref{sec:gnn-training}.

\section{Related work}\label{sec:related}

In the past GP literature, there were only few works that used a pre-trained machine learning model specifically for guiding search process. These include Neural Program Optimization \cite{Liskowski:2020:GECCO}, Neuromemetic Evolutionary Optimization \cite{10.1007/978-3-030-58112-1_43} and \cite{10.1145/3638530.3654277}. In those works, neural networks were used to model the fitness landscape spanning the solution space, in what can be broadly seen as a form of \emph{surrogate model}. \mname diverges from those approaches in several aspects. Most importantly, it does not attempt to model the fitness landscape as, for the reasons presented in Sec.\ \ref{sec:motivations}, we consider this largely futile and inconsistent withe the iterative, incremental nature of  evolutionary search. Secondly, \mname engages a GNN, while the cited approaches used other architectures of neural networks. In particular, this allows our method to learn from the detailed, fine-grained interactions between instructions of the considered DSL, rather than operate only at the level of complete programs. The mentioned works lack this kind of insight. 

To a lesser extent, \mname relates to prior works via its main constituents components, i.e. the neural network and the library of candidate solutions. In the following, we structure the review of related work accordingly.   

\mname can be seen as an evolutionary algorithm hybridized with a neural approach. Neural program synthesis has witnessed a marked acceleration in recent years due to advancements in deep learning. A notable early contribution in this area is the DeepCoder by Balog et al. \cite{2016arXiv161101989B}, where a neural network was trained to map the input-output examples provided in the program synthesis problem to the probability distribution of instructions to be used in the synthesized programs. DeepCoder utilizes this network to query the probability estimates for a given program synthesis problem. Subsequently, a search algorithm employs these estimates to parameterize its search policy, i.e. to prioritize certain instructions over others. In combination with systematic breadth-first search and other search algorithms, DeepCoder has been observed to achieve significant speedups, up to over two orders of magnitude, compared to the purely systematic search. There were also attempts to hybridize it with GP, which have been shown to boost the efficacy of evolutionary program synthesis \cite{Liskowski:2018:GECCOa}.

Several recent works demonstrated the possibility of engaging generative neural networks for 'direct' synthesis from examples. For SR, this boils down to a neural model that observes the training data and directly produces the formula as a sequence of symbolic tokens. 

Despite the fact that several architectures of this kind have been shown to achieve impressive performance on a range of benchmarks (see \cite{Biggio_Bendinelli_Neitz_Lucchi_Parascandolo_2021, Kamienny_d’Ascoli_Lample_Charton_2022}), the generative approach is subject to several limitations that have significant ramifications. These limitations are particularly pertinent to large language models (LLMs) in the context of syntactic correctness, transparency, and the ability to generalise beyond the training set. The generative approach, being essentially a sophisticated model of a conditional probability distribution, tends to interpolate between the training samples rather than extrapolate beyond them. Our method addresses these limitations by gradually constructing a formula in accordance with the adopted grammar of expressions. This approach ensures that the resulting formulas are syntactically correct by construction.  

The integration of neural inference with symbolic processing is indicative of an affiliation with the class of neurosymbolic approaches, which has recently experienced a substantial revival due to the increasing ease with which deep learning architectures can be combined with symbolic representations. Comprehensive reviews of such approaches can be found in \cite{Garcez_Lamb_2020, NSAI2022, DBLP:series/sbcs/ShakarianBSXP23}. To our best knowledge, none of such neurosymbolic approaches closely resembles \mname, in particular in its close, almost one-to-one alignment between the syntax of the analyzed programs (represented as trees extracted from the programs in the population) and the structure of the derived graph, which is subsequently processed by the GNN.

An aspect that clearly connects \mname with the prior work in evolutionary computation is its involvement of a library, which in broad terms can be seen as a form of an \emph{evolutionary archive}. The literature on archives in EC is very extensive, and its review is beyond the scope of this paper. However, in the GP field, there was significant number of past efforts aiming at forming archives of \emph{partial}, rather than complete, solutions, i.e. subprograms. Rosca and Ballard's work constitutes one of the earliest attempts of this kind \cite{rosca:1996:aigp2}. They proposed a sophisticated mechanism for assessing subroutine utility and entropy to decide when a new subroutine should be created.  As posited by Haynes (1997) employed a similar mechanism, aimed at detection of redundancy in the solutions in the population. Hornby et al. \cite{hornby_alife02_s} used  a library for the sake of reusing of assemblies of parts within the same individual. \cite{bajurnow:2004:llfegsbisp} engaged libraries with a layered learning mechanism for explicit expert-driven task decomposition. Pawlak et al. \cite{Pawlak:2014:ieeeEC} proposed to use a library of subprograms in connection with a semantic-aware search operator. On top of employing libraries for code reuse within the same evolutionary run, some researchers attempted to exploit them in separate evolutionary runs applied to other problems \cite{ryan:2004:GPTP}.

\section{Experiments}\label{sec:experiments}

The objective of the experiment is to verify and quantify the usefulness of \mname as an additional search operator for solving SR tasks. Our working hypothesis is that extending a rudimentary GP configuration with \mname improves the success rate of SR runs.

\subsection{GNN training}\label{sec:gnn-training}
For training purposes, we prepared a collection of SR problems by sampling mathematical expressions with arity ranging from 1 to 6. These expressions were built using predefined constants (0, 1, 2, 3 and $\pi$), binary operators ($+$, $-$, $\times$, $\div$) and functions ($\sqrt{x}$, $x^2$, $x^3$, $\sin$, $\cos$, $\log$, $\exp$). The complexity of each expression is determined by the total number of operators used.
For each generated expression $p$, we created a dataset $T=\{(\mathbf{x}^{(j)},y^{(j)})\}$ with $n=100$ samples. The values of independent variables $x_i$ were drawn from a normal distribution, and the corresponding outputs were computed as $y^{(j)}=p(\mathbf{x}^{(j)})$. Each pair $(p,T)$ represents a single SR problem instance.

The GNN was trained on the above described training set following the procedure outlined in Sec.\ \ref{sec:training}. Training continued until one of two stopping conditions was met: either the loss function stagnated on a validation set of 60 SR problems set aside from the training data, or the maximum limit of 1000 epochs was reached. Optimization was performed using the Adam optimizer~\cite{DBLP:journals/corr/KingmaB14} with a learning rate of \(0.001\).
Additionally, every $50$ epochs, the values of the independent variables were resampled to introduce variability in the training process.
The entire training process was performed on a single GPU and required approximately 12 hours of computation on an NVIDIA DGX machine.
It is important to highlight that this training was a one-time process: once the GNN was trained, the same model instance was used consistently across all subsequent experiments.

\subsection{Configurations of compared methods}\label{sec:method-config}

All compared configurations implement generational GP algorithm equipped with initialization, mutation, and search operators. The parameterization of \mname is almost identical to that of GP, except for the presence of the grafting operator (Sec.\ \ref{sec:grafting}). Other than that, all configurations evolve the population for 50 generations, each generation starting with evaluating solutions with the fitness function (mean square error, MSE), followed by selecting parent solutions with a tournament selection (tournament size 7), and applying search operators. We consider three configurations of search operator pipelines:
\begin{itemize}
    \item For the baseline \textbf{GP}: subtree-swapping crossover, followed by the subtree-replacing mutation applied with probability 0.2 (i.e. every fifth offspring undergoes mutation on average).
    \item \textbf{\mname}: grafting of programs from the library (Sec.\ \ref{sec:grafting}) followed by the subtree-replacing mutation configured in the same way as above. 
    \item \textbf{\mname-HH}: as \mname, however hybridized half-and-half with the subtree-swapping crossover: 50\% of selected parent solutions undergo grafting, and 50\% undergo subtree-swapping crossover, with one-fifth of the resulting programs undergoing mutation, as in the above configurations.    
\end{itemize}
To verify whether the anticipated improvements indeed originate in the guidance provided by the trained GNN, we also run experiments for an ablated version of the method, dubbed \textbf{\mname-Abl}. This setup mimics the \mname configuration, except for the selection step of the expander (Sec.\ \ref{sec:expander}): here, rather than relying on the GNN, we pick the expanded graph nodes at random (up to 5 of them, as in \mname). 

Offspring solutions that exceed height 13 are discarded and replaced by their parents. The outcome of a run is the solution with the lowest MSE found throughout the run. If the MSE is smaller $10^{-10}$, we deem the run as successful.

The mutation operator uniformly draws a node in the parent tree and replaces the subtree rooted in that node with a subtree generated as follows: first, $h'$ is drawn uniformly from the $[0,h]$ interval, then a random tree of height $h'$ is generated using the `grow' method and grafted at the selected node. 

We perform experiments for population sizes 100, 200, 500, and $1{,}000$. Recall that population size determines also the capacity of the library (Sec.\ \ref{sec:library}).  

We implemented \mname using using the DEAP library \cite{DEAP_JMLR2012}.

\subsection{Test suite}\label{sec:test-set}

Our set of test problems is the AI Feynman collection of regression problems \cite{Udrescu_Tegmark_2020}. This suite comprises 100 equations selected from the \emph{Feynman Lecture on Physics}. Of those, three problems use the $\arcsin$ and $\tanh$ functions, which are not present in our instruction set. We therefore discarded those equations, and thus the resulting test set comprises 97 instances of SR problems.  

A single configuration considered in this experiment (i.e. a method setup combined with a population size) involved thus 97 evolutionary runs (over $1{,}500$ runs in total). 

\subsection{Results}\label{sec:results}

In Table \ref{tab:success-rate}, we summarize the success rates of compared configurations, obtained on the test set of 97 problems outlined above. The figures are clearly better for \mname, which clearly indicates that the guidance provided by the GNN is informative, and that the proposed informed search operator is capable of predicting the prospective utility of expressions constructed in the expansion process.  

\begin{table}[t]
  \caption{Success rates (percentage of successful runs out of the 97 test problems) for the runs using different population size. 
  }\label{tab:success-rate}
  \begin{tabular}{lcccc}
    \toprule
    Configuration   & \multicolumn{4}{c}{Population size} \\
                 \cmidrule(lr){2-5}
    & 100 & 200 & 500 & 1000 \\
    \hline
    GP & 0.0722	&	0.1237 &		0.1753	&	0.2062 \\
    \mname &   0.1340 &	0.1443	&	0.1959	&	0.1959 \\
    \mname-HH &  0.1031 &		0.1546	&	0.2062 &		0.2062 \\
    \midrule
    \mname-Abl & 0.0722	&	0.1031	&	0.1856 & 0.2268 \\
    \bottomrule
\end{tabular}
\end{table}

This is corroborated by the observation that the success rates for \mname and \mname-HH tend to be also better than for the ablated variant, \mname-Abl. The only exception is the case of population size $1{,}000$, where \mname-Abl outperforms all remaining configurations. Recall that the capacity of the library is the same as the population size. This suggests that a large library, supplied with a diversified sample of subprograms extracted at random (though with a bias towards larger subtrees) can be also helpful at making search more efficient. Nevertheless, this specific result for population size $1{,}000$ does not invalidate the trends observed for the remaining population sizes. 

The \mname-HH variant of the proposed method tends to be slightly better than \mname for population sizes 200 and larger, which suggests that partial reliance on the conventional subtree-swapping crossover may bring some benefits. However, this advantage is not present for the smaller considered population sizes. This may indicate that when the `memory' of the search process, embodied by the population and the library, is compact, it may be desirable to rely stronger on the informed guidance provided by the GNN.  

\begin{table}[t]
  \caption{Success rates for \mname variants, factored by problem arity (number of input variables). No configuration managed to solve problems of arity 6 or higher. 
  }\label{tab:success-arity}
  \begin{tabular}{lccccc}
    \toprule
    Configuration   & Arity &  \multicolumn{4}{c}{Population size} \\
                 \cmidrule(lr){3-6}
    & & 100 & 200 & 500 & 1000 \\
    \toprule
    GP & 2 & 0.3571	& 0.4286	& 0.4286	& 0.5000 \\
    GP & 3 & 0.0286	& 0.0571	& 0.1714	& 0.2286 \\
    GP & 4 & 0.0370	& 0.1481	& 0.1852	& 0.1852 \\
    \midrule
    \mname & 2 & 0.3571	& 0.4286	& 0.5714	& 0.5714 \\
    \mname & 3 & 0.1143	& 0.1143	& 0.1714	& 0.1429 \\
    \mname & 4 & 0.1481	& 0.1481	& 0.1852	& 0.2222 \\
    \midrule
    \mname-HH & 2 & 0.3571	& 0.3571	& 0.5000	& 0.5714 \\
    \mname-HH & 3 & 0.0571	& 0.0857	& 0.2000	& 0.2000 \\
    \mname-HH & 4 & 0.1111	& 0.0741	& 0.2308	& 0.1481 \\
    \mname-HH & 5 & 0.0	& 0.0	& 0.0	& 0.0833 \\
    \bottomrule
\end{tabular}
\end{table}

For all compared methods, the success rates are admittedly relatively low in absolute terms. This is mostly due to the high level of difficulty of SR instances in the AI Feynman suite \cite{Udrescu_Tegmark_2020}. One of the sources of that difficulty is the relative high arity of expressions, which involve up to 9 independent variables (inputs). To provide a better insight in this aspect, in Table \ref{tab:success-arity} we present the success rates factored by the problem arity. In the test set, there are 14, 35, 12 and 27 problem with arity 2, 3, 4 and 5, respectively. None of the methods solved instances with more than 5 inputs.\footnote{In the AI Feynman suite, there is only one problem instance with arity 1, so we dropped it from this table for clarity.} \mname-HH is the only configuration which, when ran with population of size $1{,}000$, has managed to solve a single instance of 5-variable problem.

Table \ref{tab:size} presents the sizes of successfully synthesized programs (i.e. the number of tree nodes), factored by problem arity. Also on this metric \mname fares better than GP: the expressions it synthesized are much smaller (often several times) than those produced by GP. We may thus conclude that the GNN-based guidance does not only helps search to converge on correct solutions, but also avoids redundancy in construction of candidate solutions. 

Inspection of sizes of successfully synthesized solutions leads to another interesting observation. Recall that we bias the expander so that it tends to pick the larger subtrees $s$ for expansion, rather than the smaller ones (Sec.\ \ref{sec:expander}). Despite this, the trees synthesized by \mname and \mname-HH are small. This suggest that it is probably quite common for the GNN to select for the library not the expressions created by combining the entire $s$ with other arguments using one of the operators, but by combining \emph{one of the subexpressions of $s$} (which also participate in the expansion process; see Fig.\ \ref{fig:graph-example}).

\begin{table}[t]
  \caption{Sizes of successfully synthesized expressions (in nodes), factored by problem arity.}\label{tab:size}
  \begin{tabular}{lccccc}
    \toprule
    Configuration   & Arity &  \multicolumn{4}{c}{Population size} \\
                 \cmidrule(lr){3-6}
    & & 100 & 200 & 500 & 1000 \\
    \toprule
    GP & 2 & 9.4 & 24.2 & 15.3 & 22.4 \\
    GP & 3 & 8.0 & 9.0 & 39.7 & 53.5 \\
    GP & 4 & 7.0 & 21.2 & 25.4 & 42.8 \\
                 \cmidrule(lr){2-6}
    & Avg. & 8.6 & 20.7 & 26.9 & 40.2 \\
    \midrule
    \mname & 2 & 5.4 & 7.2 & 5.7 & 5.2 \\
    \mname & 3 & 7.0 & 6.7 & 9.5 & 6.6 \\
    \mname & 4 & 8.0 & 7.25 & 7.8 & 7.8 \\
                 \cmidrule(lr){2-6}
    & Avg. & 6.7 & 7.1 & 7.5 & 6.42 \\
    \midrule
    \mname-HH & 2 & 3.8 & 3.4 & 4.9 & 5.7 \\
    \mname-HH & 3 & 5.0 & 11.0 & 7.1 & 7.3 \\
    \mname-HH & 4 & 8.3 & 7.0 & 9.5 & 7.2 \\
    \mname-HH & 5 & -- & -- & -- & 9.0 \\
                 \cmidrule(lr){2-6}
    & Avg. & 5.4 & 7.9 & 7.0 & 6.7 \\
    \bottomrule
\end{tabular}
\end{table}

\section{Discussion}\label{sec:discuss}

The observed improvements in success rates are not very large. This is in part due to the high level of difficulty of the AI Feynman suite, which features equations capturing complex real-world physical laws (rather than, e.g., expressions generated at random). We also hypothesize that another reason is the difficulty of the guidance task faced by the GNN. It is worth realizing that the GNN is meant here to mimic the process of directed, thoughtful scientific discovery: given the currently available expressions (materialized in the graph), the values they produce for each training example $\mathbf{x}^{(j)}$, and the desired values of the dependent variable (the target $y^{(j)}$), the GNN is expected to indicate which of those expression are good candidates for becoming a part of the final regression model. In that process, the GNN should relate the syntactic information about program instructions to their semantics (the effect those instructions have on data processing), which is known to be very hard in general. 

In connection with the motivations in Sec.\ \ref{sec:motivations}, we find it particular important that \mname is capable of conducting informative search guidance in the space of \emph{partial solutions}, rather than complete programs. As argued there, it is in general not reasonable to assume that the fitness function is capable of anticipating the prospective usefulness of a solution component (here: a subtree). As shown in many past studies, the fitness landscape in GP, and in particular in symbolic regression, is extremely rugged, so that even the programs that are syntactically very similar to the correct one, differing from it by just one instruction, can have very low fitness. We posit further qualitative progress in the EC field is conditioned on designing such better-informed search operators.

\section{Conclusions}\label{sec:conclusions}

The presented results constitute preliminary evidence for the usefulness of machine learning, and in particular deep learning-based GNNs, as a means of designing informed search operators for GP, and in particular for SR. On average, \mname increases the likelihood of constructing correct solutions, which are also much smaller than those produced by the baseline GP. 

While in this paper we have demonstrated \mname in the domain of SR, it is worth emphasizing that it can be quite easily ported to other domains in which candidate solutions can be represented as trees or graphs, and such in which one can naturally define an expansion step. Interestingly, this includes also domains beyond program synthesis: while the GNN in \mname observes the semantic effects of expression execution, this is not an indispensable component of this approach.

\begin{acks}
  This research was supported by the statutory funds of Poznan University of Technology and the Polish Ministry of Education and Science, grant no. 0311/SBAD/0752, and by National Science Centre, Poland, grant no. 2024/53/N/ST6/03961.
\end{acks}